\documentclass{article}
\usepackage{spconf,amsmath,graphicx}
\usepackage{tabularx,booktabs}
\usepackage{subcaption}

\DeclareMathOperator*{\argmax}{argmax}

\usepackage{relsize}

\title{A differentiable Gaussian Prototype Layer for explainable Segmentation}

\name{Michael Gerstenberger, Steffen Maaß, Peter Eisert, and Sebastian Bosse
\thanks{This research has received funding from the German Federal Ministry for Economic Affairs and Climate Action as part of the NaLamKI project under Grant 01MK21003D}
}

\address{Fraunhofer HHI, Berlin, Germany}
\begin{document}
%
\maketitle
\begin{abstract}
 We introduce a Gaussian Prototype Layer for gradient-based prototype learning and demonstrate two novel network architectures for explainable segmentation one of which relies on region proposals. Both models are evaluated on agricultural datasets. While Gaussian Mixture Models
(GMMs) have been used to model latent distributions of neural networks before, they are typically fitted using the EM
algorithm. Instead, the proposed prototype layer relies on gradient-based optimization and hence allows for end-to-end training. This facilitates development and allows to use the full potential of a trainable deep feature extractor. We show that it can be used as a novel building block for explainable neural networks. We employ our Gaussian Prototype Layer in (1) a model where prototypes are detected in the latent grid and (2) a model inspired by Fast-RCNN with SLIC superpixels as region proposals. The earlier achieves a similar performance as compared to the state-of-the art while the latter has the benefit of a more precise prototype localization that comes at the cost of slightly lower accuracies. By introducing a gradient-based GMM layer we combine the benefits of end-to-end training with the simplicity and theoretical foundation of GMMs which will allow to adapt existing semi-supervised learning strategies for prototypical part models in future.

\end{abstract}
\begin{keywords}
Prototype-based learning, fruit segmentation, Gaussian Mixture Models (GMMs), region proposals
\end{keywords}

\section{Introduction}
\label{sec:intro}

Prototype-based learning combines deep feature extraction with the detection of prototypes in latent space and promises better explainable predictions as compared to conventional networks \cite{chen2019looks}. Predictions are computed by matching features of an image and learned prototypes that capture decisive image elements and then using the prototype activation of a class to make a decision. Prototype-based deep learning relies on the idea of identifying the similarity between a given sample and known reference points. It relates to the scheme of explaining by examples and is rooted in cognitive science\cite{gerstenberger2022but}.\\
Most commonly prototype networks are used to classify images and identify charactersitic visual features of the classes \cite{chen2019looks}. Differences exist regarding the way prototypes are fitted, using iterative \cite{chong2021towards} or gradient-based schemes with prototype pruning \cite{chen2019looks} or without \cite{rymarczyk2022interpretable}, the question if negative reasoning is allowed \cite{stefenon2022semi, singh2021these} or prohibited \cite{chen2019looks} the backbone (CNNs \cite{chen2019looks, chong2021towards, rymarczyk2022interpretable}, GNNs \cite{zhang2022protgnn} or transformers\cite{kim2022vit}) and whether relationships between prototypes such as taxonomies \cite{hase2019interpretable} or configurations in pixel space are modeled \cite{donnelly2022deformable}. Besides, prototype methods have been developed for the segmentation of videos and to enable few shot learning with support sets \cite{dong_few-shot_2018, ke2021prototypical}.\\
Here, we extend on the work on explainable classification with positive-only reasoning by \cite{chen2019looks} and show that prototype based learning can help in understanding how segmentations are computed. Until now prototype architectures that employed GMMs have been trained iteratively using expectation maximization \cite{ke2021prototypical}. We considered previous work on gradient-based optimization of GMMs for high dimensional data \cite{gepperth2021gradient} and demonstrate a novel type of prototype layer that allows for end-to-end training. Moreover, we present use-cases of this layer in two novel networks: In ProtoSegNet prototypes are detected in the latent grid retrieved using a ResNet encoder and in ProtoBBNet in embedded region proposals. They are used to tackle the challenge of a precise prototype localization that has only been addressed using LRP before \cite{gautam2023looks}. As our major contribution we show that a gradient-based Gaussian Prototype Layer can serve as a building block in networks for explainable segmentation and in combination with region proposals. It is introduced in section 2. The models are explained in the subsections 2.1 and 2.2 respectively. An evaluation and the comparison to the state-of-the-art for fruit segmentation follows in section 3. 



\section{Methods}
 Our prototype layer builds on the idea of modeling latent distributions using class specific GMMs. Each 
 class has a set of gaussian kernels that are associated with prototypical image regions. The conditional probabilities per class allow to classify latent embeddings. ProtoSegNet detects prototypical features directly in the latent grid while ProtoBBNet further encodes the grid vectors suggested by a region proposal mechanism and then detects prototyes for classification. \\
We developed a Gaussian Prototype Layer as it has several desirable properties. Similar to prototypical part models for classification it explains segmentations with a "this-looks-like-that" rule and allows to identify prototypical image regions (Fig. 1). As it is end-to-end trainable it is well suited for transfer learning and we observe higher accuracies as compared to a pretrained feature extractor with fixed weights (75\% accuracy as compared to 99\% for MinneApple). Anistoropic kernels also promise a better fit with fewer kernels especially if the encoder is not trained which is relevant for realtime user interaction. Moreover, as latent space is modeled by a GMM, existing interactive learning strategies such as those for semi-supervised learning\cite{xiong2010semi} can also easily be used with our Gaussian Prototype Layer.
\subsection{ProtoSegNet: Grid Prototypes for Segmentation}
ProtoSegNet operates directly on the ResNet embeddings of an image. The grid has a shape of $n^\prime \times n^\prime \times l$ where $n^\prime$ stands for the width and height and $l$ refers to the number of feature maps. Prototypes are detected in the space, spanned by the grid vectors (Fig 1a). The approximate position of a detected prototype in pixel space can be inferred from the position of the grid vector which is most similar to a given prototype for the dataset in question using a simple upscaling rule \cite{chen2019looks}. This allows to identify prototypical image regions of the training set. Conceptually, a prototype consists of the prototypical image-patch, its latent encoding and the corresponding closest prototype vector \cite{chen2019looks} which is the mean of the assigned Gaussian mixture component in the presented approach. Moreover each prototype codes for a specific class that is defined a priory (here: fruit and background). \\
Let $x\in\mathbf{R}^{3\times n\times n}$ be a normalized RGB-image of width and
height $n$. The convolutional embedding is a function $f_\theta$ that maps
the image into the latent space 
where $\ell$ is the number of channels in the latent space and $n'$ the
width and height of the feature map in the latent space.
    \begin{align*}
    z_{i,j=1,\dots,n'} 
    = f_\theta(x)\in \left(\mathbf{R}^{\ell}\right)^{n'\times n'}
\end{align*}
The spatial scaling factor
of the convolutional embedding is then $p = n/n'$.
Each latent vector 
$z_{i,j}\in\mathbf{R}^\ell$ corresponding to the image patch is then classified independently from the other latent vectors in the feature map.
\begin{align*}
    (x_{i p+\alpha, j p+\beta})_{\alpha,\beta=1,\dots,p}
    \in\mathbf{R}^{p\times p}
\end{align*}

A GMM is used to model the distribution of latent embeddings where each mixture component corresponds to a prototype and its mean $\mu$ represents the prototype vector. Each component has a covariance matrix $\Sigma$ and a parameter $w$ of the distibution  associated with component $i$.  
\begin{align*}
    p = \sum_{k=1}^Nw_k \mathcal{N}(\mu_k,\Sigma_k), \text{where}\\
    Z | K \sim \mathcal{N}(\mu_K, \Sigma_K)
\end{align*}
with $\mu_1, \dots, \mu_N\in\mathbf{R}^\ell$ and $\Sigma_1, \dots,\Sigma_N
\in\mathbf{R}^{\ell\times\ell}$ and $K\in\{1, 2, \dots, N\}$.
For the conditional distribution of $K$ given $Z=z\in\mathbf{R}^\ell$ then holds:

\begin{align*}
\resizebox{.99\hsize}{!}{$
    \log p(K| Z = z) \propto w_K - \frac{n}{2}\log\left(2\pi|\Sigma_K|\right)
    -\frac{1}{2}(z-\mu_K)^T \Sigma_K^{-1}(z-\mu_K).
    $}
\end{align*}
The final layer is linear, constrained to positive weights, has no bias and yields for each class, $y\in\{1,2,\dots,C\}$ a class score $c_y$.
\begin{align*}
    c_y = \sum_{k=1}^N a_{y, k} \log p(K=k|Z = z), \text{where}\\
    A = (a_{y, k})_{y=1,\dots,C; k=1,\dots,N}\in\mathbf{R}_{\geq 0}^{C\times N}
\end{align*}
    
Consequently the classification of an image patch $x$ from an image 
$x\in\mathbf{R}^{3\times n\times n}$ is given as follows.
\begin{align*}
    y = \argmax\limits_{y=1,\dots,C}~~\sum_{k=1}^N a_{y, k} 
    \log p\left(K=k|Z = f_\theta(x)_{i,j}\right).
\end{align*}
 Hence, the class of each ground-truth annotation is computed from a linear combination of the logarithmically-scaled probability masses. After convergence of the loss, the prediction of the output for a class is a linear combination of the scaled probability masses of the class prototypes.\\
The model is trained in four stages. The convolutional embedding 
is the composition of $m$-ResNet blocks (with $m\in\{1,2,3,4\}$), $g_{\theta'}$, 
and $1\times 1$-convolution layers, $h_{\theta''}$ that reduce the number of channels.
The channel reduction is first trained via the autoencoder 
$H_\phi\circ h_{\theta''}$ using the MSE loss. Next the GMM is fitted on the training data.
Let 
$
    z^{(1)} = f_\theta(x^{(1)}), \dots, z^{(d)} = f_\theta(x^{(d)})
$
be the embedded images and denote by $\mathbf{\eta}$ the parameters
of the GMM. 
\begin{align*}
    L_{\text{GMM}}(\mathbf{\eta}; z^{(1)}, \dots, z^{(d)}) = 
    -\log\mathcal{L}(\mathbf{\eta}; z^{(1)}, \dots, z^{(d)})
\end{align*}
Then the negative log-likelihood function $L_{\text{GMM}}$ is minimized. The weights of the final linear layer are trained in a supervised way using the crossentropy $H$:
\begin{align*}
    &L_{\text{clf}}
    &= \frac{1}{d (n')^2}\sum_{\alpha=1}^d\sum_{i,j=1}^{n'} 
    H\left(A \log p\left(K|Z = z^{(\alpha)}_{i,j}\right), y^{(\alpha)}_{i,j}\right)
\end{align*}
Finally, all parameters 
$\theta, w, \mu, \Sigma$ and $ A$
are trained jointly with respect to the combined loss function $L$, where L1 is a regularization term that minimizes the weights between the output neuron of a class and prototypes of other classes.
\begin{align*}
     L = L_{\text{GMM}} + L_{\text{clf}} + L1
\end{align*}
\subsection{ProtoBBNet: Prototypes in Region Proposals}
In ProtoSegNet the location and outline of prototypical patches can only be approximated using the upscaling rule \cite{gautam2023looks}. To facilitate future user interaction we hence assess a different network architecture for ProtoBBNet. It is inspired by Fast-RCNN but uses SLIC superpixels for the region proposal \cite{girshick2015fast}. The model employs prototype detection in the space spanned by the embedded region proposals obtained by cropping and encoding areas of the latent grid that correspond to the identified SLIC superpixels \cite{achanta2010slic}. In ProtoBBNet, segmentation is achieved by assigning the predicted class of the embedding to the area of the SLIC-superpixels in the images.\\
Our second model comprises of a region proposal mechanism, a deep encoder (ResNet) and a secondary encoder with four convolutional layers, batch normalization and leaky-RELU activation function (Fig 1). First the image embedding is obtained. Second, we retrieve region proposals by segmenting the image into $m$ SLIC superpixels and obtaining the enclosing bounding box for each segment. Each region proposal is labeled as foreground if the majority of ground-truth pixels of the superpixel indicates the respective class. Otherwise it is labeled as background. These bounding boxes are used to crop the corresponding area of the latent-grid using RoIAlign \cite{he2017mask} and the resulting volume is further encoded to yield exactly one latent vector per region proposal. The latent vectors $z'$ are encoded by the secondary encoder $f_\phi$ which takes the aligned patches cropped from the output of the primary encoder $f_\theta$.

    \begin{align*}
        z'_{i\dots,m'} 
        = f_\phi(\textsf{RoiAlign}(f_\theta(x), \textsf{Roi(x)}))\in \left(\mathbf{R}^{\ell}\right)^{1\times 1}
    \end{align*}

Analogously to ProtoSegNet, the logarithmically-scaled mass function of the prototypes is then evaluated given the latent space spanned by $z'$ and a linear combination computed to predict the class of each superpixel.  

The training of ProtoBBNet comprises of four stages. First, an autoencoder is used to train the secondary encoder that embeds the region proposals. Second, the prototypes are fitted in latent space by training the GMM on the latent distribution. Third, the classification head is trained while the final and fourth stage comprises of end-to-end training.

\section{Results}
\label{sec:typestyle}

\begin{figure}[ht]
\begin{subfigure}[b]{0.5\textwidth}
   \includegraphics[width=1\linewidth]
   {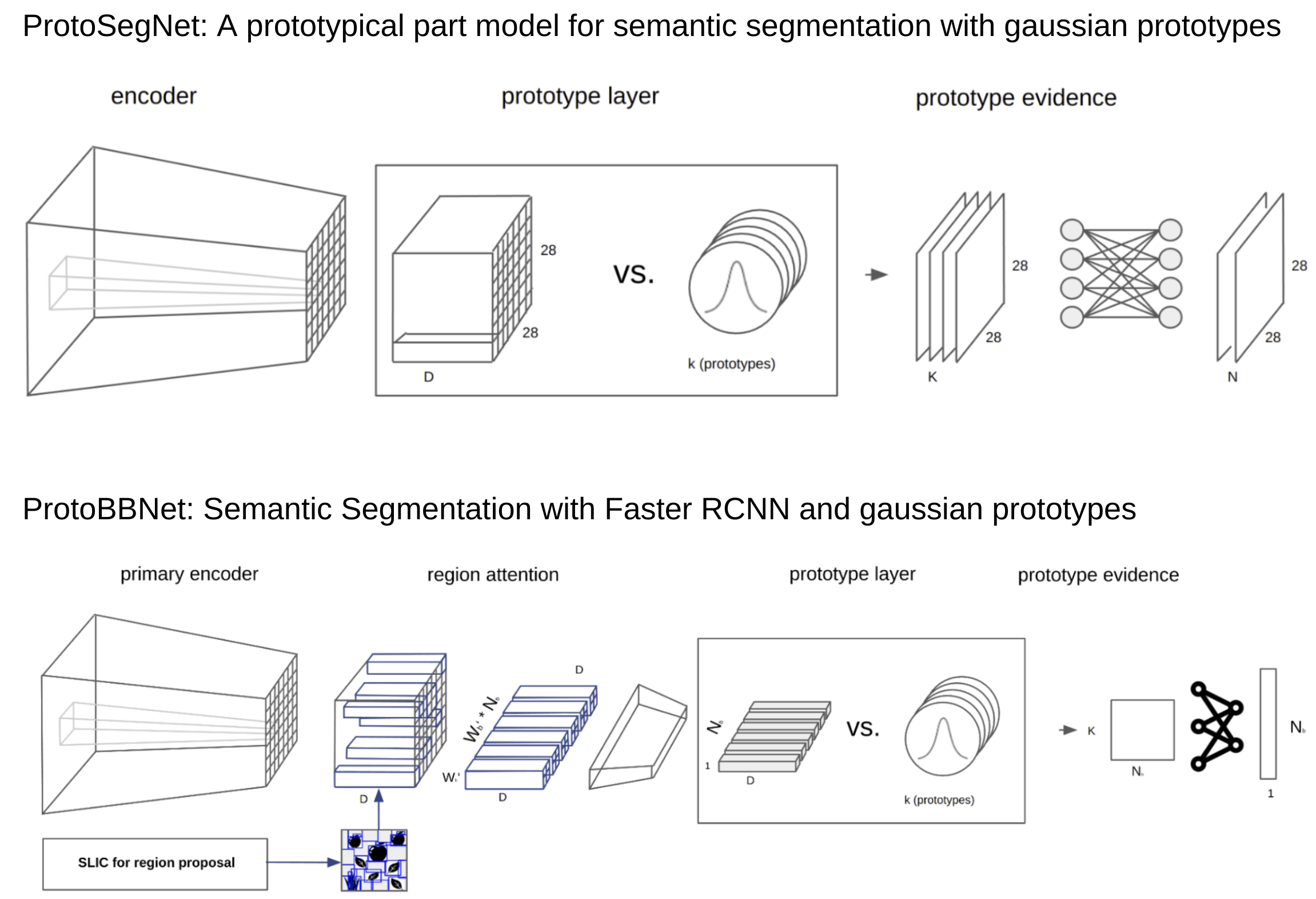}
   \caption{Network Architectures}
   \label{fig:Ng1} 
\end{subfigure}

\begin{subfigure}[b]{0.5\textwidth}
   \includegraphics[width=1\linewidth]
      {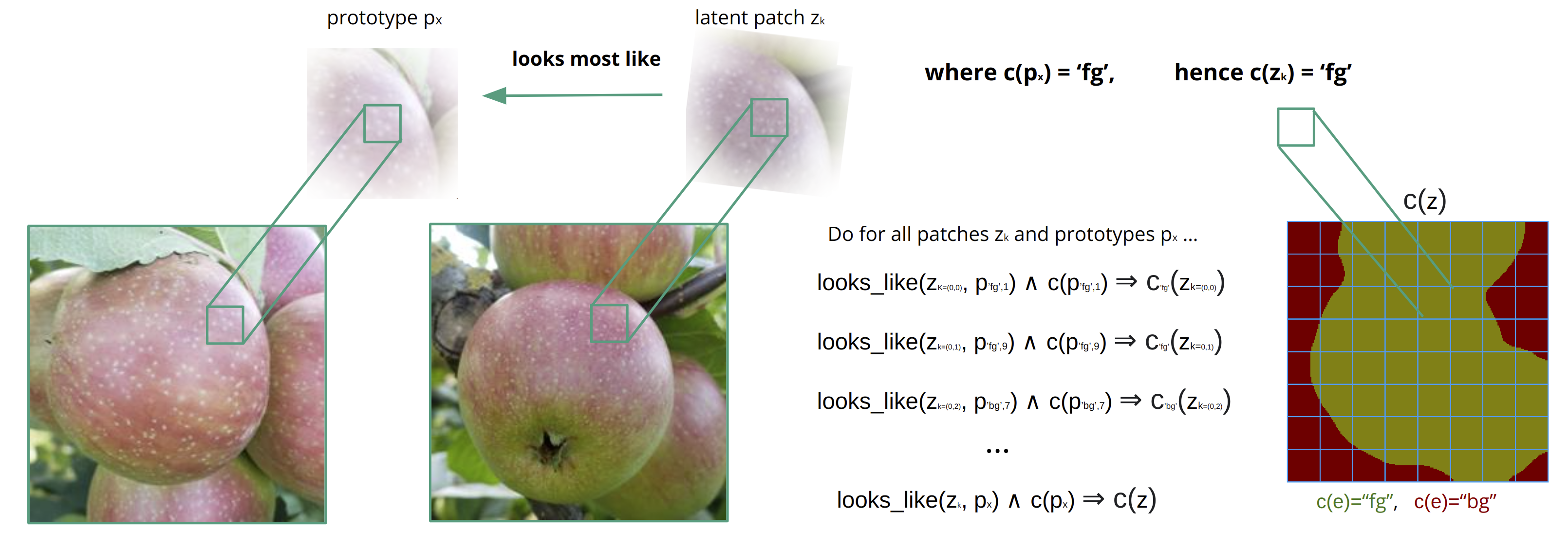}

   \caption{The logic of inference}
   \label{fig:Ng2}
\end{subfigure}
\caption{Prototype based segmentation}
\end{figure}

\begin{table*}
\scalebox{0.65}{
\begin{tabularx}{1.5\textwidth}{p{30mm}p{50mm}p{20mm}p{20mm}p{20mm}p{30mm}p{30mm}p{30mm}}
\toprule
Dataset & Approach     & Explainable & Backbone & Mean IoU & Class IoU & Pixel Accuracy & Class Accuracy\\ 
\midrule
MinneApple&Colorspace GMM (Superpixels)     & no & -        & 0.512   & 0.167  & 0.862 &  0.776 \\
&U-Net 50 (no pretraining)   & no & ResNet50 & 0.678 & 0.397 & 0.960 & 0.818 \\
&U-Net 18 (no pretraining)   & no & ResNet18 & 0.691 & 0.405 & 0.978 & 0.709 \\
&U-Net 50 (ImageNet pretraining) & no & ResNet50 & 0.685 & 0.410 & 0.962 & 0.848\\
&U-Net 18 (ImageNet pretraining)   & no & ResNet18 & 0.683 & 0.393 & 0.974 & 0.590 \\
&ProtoSegNet (ours) & yes & ResNet18 & 0.678 & 0.393 & 0.963 & 0.941\\
&ProtoBBNet (ours) & yes & ResNet18 &0.708 &0.344 &0.975 &0.423 \\
\bottomrule
WGISD Grapes&Colorspace GMM (Superpixels)     & no & -        & 0.512   & 0.292  & 0.759 &  0.780 \\
&U-Net 50 (no pretraining) & no & ResNet50 & 0.721 & 0.512 & 0.935 & 0.928\\
&U-Net 18 (no pretraining)  & no & ResNet18 & 0.711 & 0.493 & 0.934 & 0.946\\
&U-Net 50 (ImageNet pretraining) & no & ResNet50 & 0.797 & 0.648 & 0.951 & 0.890\\
&U-Net 18 (ImageNet pretraining)& no & ResNet18 & 0.788 & 0.638 & 0.944 & 0.787\\
&ProtoSegNet (ours) & yes & ResNet18 & 0.765 & 0.609 & 0.930 & 0.845\\
&ProtoBBNet (ours) & yes & ResNet18 & 0.794 &0.515 &0.953 &0.603\\
\bottomrule
\end{tabularx}}
\label{tab:results}
\end{table*}
We evaluate our explainable segmentation models on two agricultural datasets: MinneApple that contains semantic segmentation of apples in an orchard and WGISD with segmented grapes\cite{hani2019minneapple,santos2020grape}. We compare them to other state-of-the-art approaches for semantic fruit segmentation: An approach that classifies superpixels based on their color as well as variations of UNet with ResNet feature extractors \cite{hani2020minneapple}. Results are summarized in Table 1.  \\
In the first comparison method, the average HSV values are extracted for each detected superpixel. The result is fed into a (semi-) supervised GMM that models the distribution of each class with independent kernels and predicts the label of a superpixel by evaluating the relative probability density \cite{xiong2010semi}. To ensure the best possible fit for the comparison we used fully supervised learning. The accuracy for MinneApple is 86\% while 77\% is achieved for WGISD based on the color GMM. The class accuracies for apples and grapes respectively are slightly lower at 78\% for both datasets. \\
The second comparison method is UNet with ResNet backbone. We use color images and compare encoder pretraining on ImageNet to training from scratch. The best pixel accuracy for MinneApple is 98\% (UNet18) which is slightly higher than 95\% accuracy for WGISD. The best class accuracies for UNet are 85\% (MinneApple) and 95\% (WGISD). \\
While we train the model directly on the WGISD data in the case of MinneApple we divide the images into patches first as training on the full-sized images requires a large amount of memory. An adequate number of training epochs is chosen such that the GMM and classification terms of the loss function converge. For the region proposal of ProtoBBNet, we use 500 regions for the cropped MinneApple images and 1000 for WGISD. ProtoSegNet achieves 96\% accuracy on MinneApple and 93\% on the WGISD dataset. Class accuracies for the fruits are 94\% and 85\%. Both indicate a performance that is at most 2\% off the best UNet results. While the class accuracies are lower (42\% and 60\%) ProtoBBNet achieves a slightly better pixel accuracy 98\% and 95\%.
To identify prototypical image regions we find the mixture component with the highest conditional probability, and use the upscaling rule (ProtoSegNet) or superpixel boundaries (ProtoBBNet).\\
\begin{figure}[ht]
\begin{subfigure}[b]{0.5\textwidth}
   \includegraphics[width=1\linewidth]
   {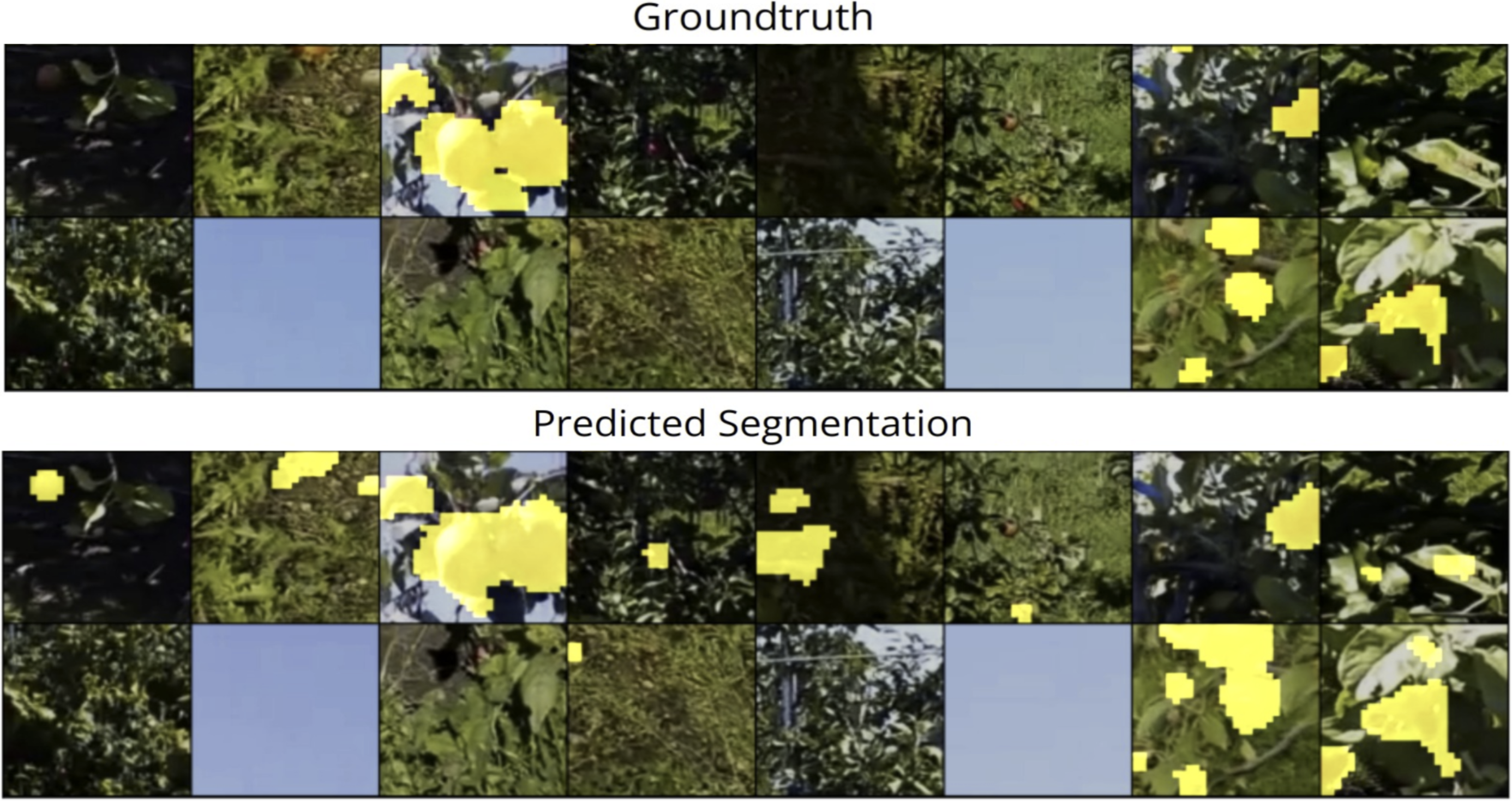}
   \caption{Segmentation results}
   \label{fig:} 
\end{subfigure}

\begin{subfigure}[b]{0.5\textwidth}
   \includegraphics[width=1\linewidth]
      {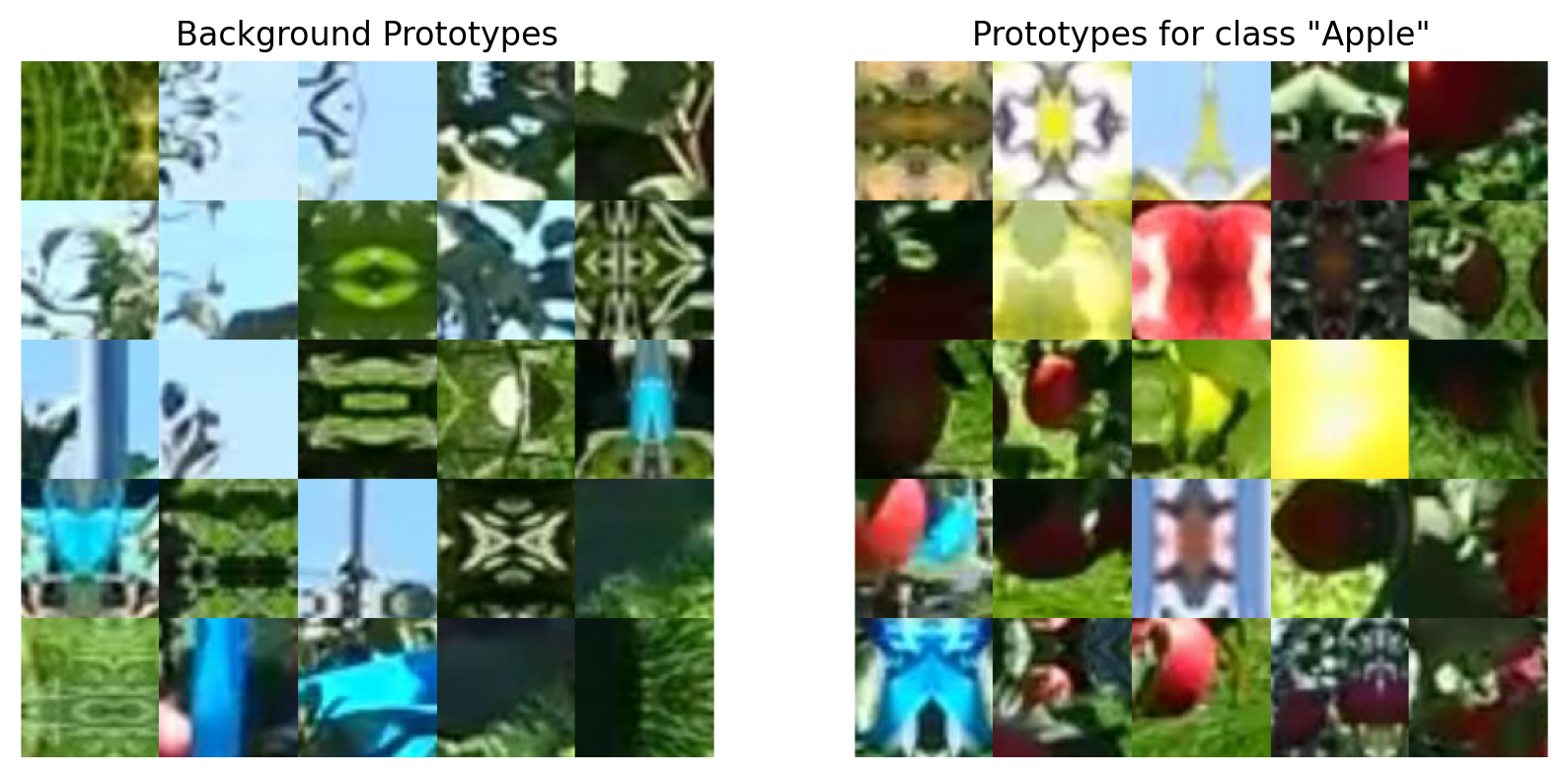}

   \caption{Prototypical image patches}
   \label{fig:Ng2}
\end{subfigure}
\caption{Segmentation of the MinneApple dataset with a prototypical parts model for segmentation (ProtoSegNet).}
\end{figure}
\begin{figure}[h!]
\includegraphics[width=\linewidth]
{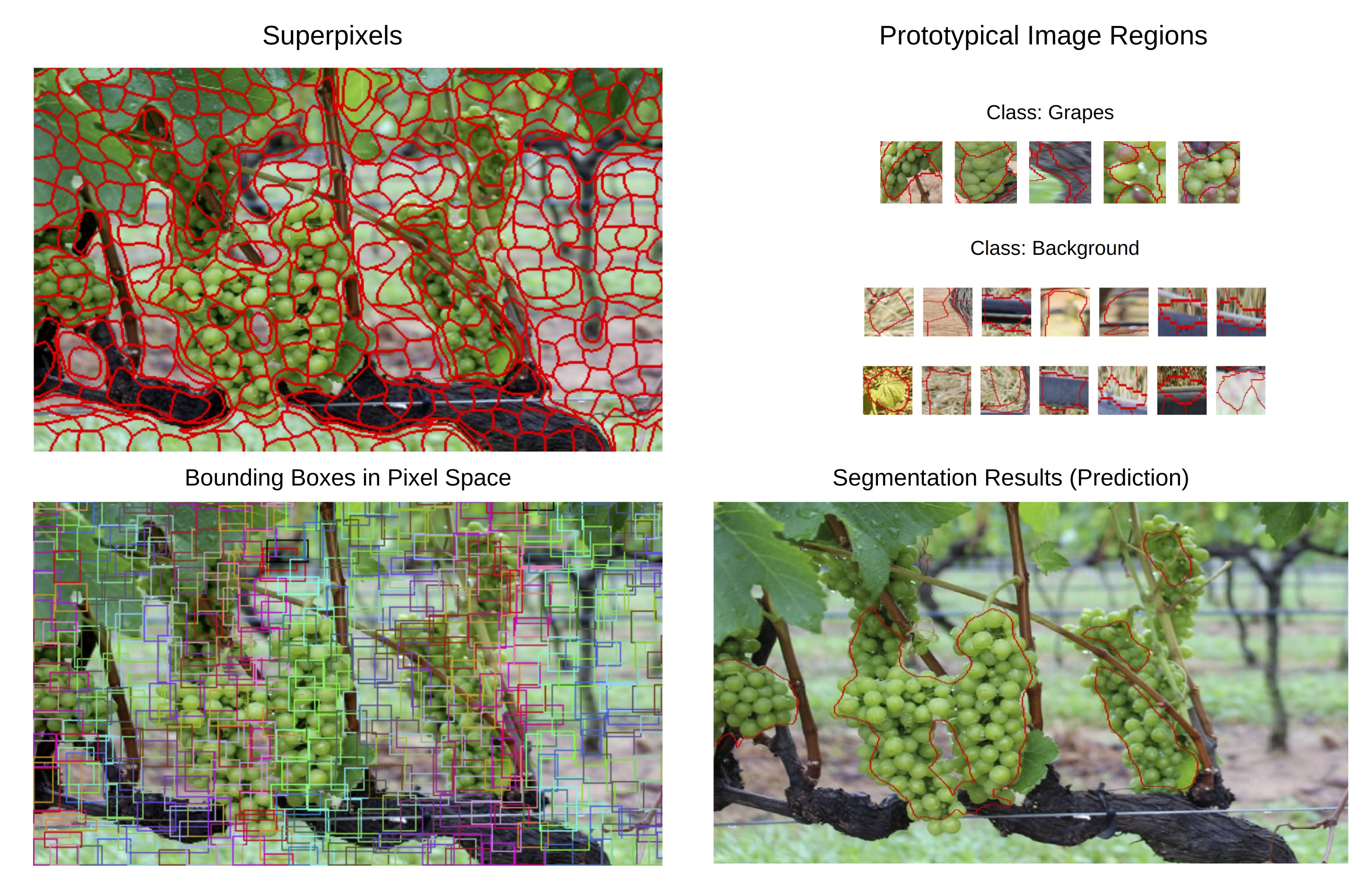}
        \caption{Segmentation of WGISD with a Fast-RCNN architecture and Gaussian Prototypes (ProtoBBNet)}
\end{figure}

The models do not only compute segmentations but also explain it. The prototypes reveal the characteristic features of the classes. Fig. 2b shows prototypical image regions for ProtoSegNet fitted on the MinneApple dataset. The foreground prototypes represent a visually heterogeneous set of apples while the background prototypes are mostly sensitive to leaves and the sky. Similarly, we identify the proposed region with an embedding that has the highest conditional probability for each prototype in ProtoBBNet. The respective superpixel show prototypical image patches of both classes. Even though other image regions (e.g. those showing leaves) are labeled as background too, the model identified patches of dry grass as being most characteristic for the class background. This reflects the understanding of prototypes to code for features that are central to the class (See Fig. 3, top right).  

\section{Summary and Outlook}
\label{sec:print}
We tested two novel network structures that employ an end-to-end-trainable Gaussian prototype layer. The models do not only perform semantic segmentation but also reveal the prototypes that explains it. ProtoSegNet achieved segmentation accuracies in the range of the state-of-the-art for the given datasets (98\% and 95\% pixel accuracy) which is slightly better then ProtoBBNet (96\% and 93\%). ProtoBBNet has the benefit of a superpixel-based region proposal which allows to attribute image regions with precise outlines to the prototypes and latent vectors. Hence it is a good candidate for the development of interactive segmentation: A user could interactively define the parameters of SLIC, inject superpixels to be used as prototypes and improve segmentations by selecting/deselecting superpixels classified by the model. More traditional region proposal mechanisms such as Selective Search could also enable interactive object localization while both models can also easily be modified for image classification.
\newpage


\bibliographystyle{IEEEbib}
\bibliography{paper}

\begin{thebibliography}{10}

\bibitem{chen2019looks}
Chaofan Chen, Oscar Li, Daniel Tao, Alina Barnett, Cynthia Rudin, and
  Jonathan~K Su,
\newblock ``This looks like that: deep learning for interpretable image
  recognition,''
\newblock {\em Advances in neural information processing systems}, vol. 32,
  2019.

\bibitem{gerstenberger2022but}
Michael Gerstenberger, Sebastian Lapuschkin, Peter Eisert, and Sebastian Bosse,
\newblock ``But that's not why: Inference adjustment by interactive prototype
  deselection,''
\newblock {\em arXiv preprint arXiv:2203.10087}, 2022.

\bibitem{chong2021towards}
Penny Chong, Ngai-Man Cheung, Yuval Elovici, and Alexander Binder,
\newblock ``Towards scalable and unified example-based explanation and outlier
  detection,''
\newblock {\em IEEE Transactions on Image Processing}, 2021.

\bibitem{rymarczyk2022interpretable}
Dawid Rymarczyk, {\L}ukasz Struski, Micha{\l} G{\'o}rszczak, Koryna
  Lewandowska, Jacek Tabor, and Bartosz Zieli{\'n}ski,
\newblock ``Interpretable image classification with differentiable prototypes
  assignment,''
\newblock in {\em Computer Vision--ECCV 2022: 17th European Conference, Tel
  Aviv, Israel, October 23--27, 2022, Proceedings, Part XII}. Springer, 2022,
  pp. 351--368.

\bibitem{stefenon2022semi}
Stefano~Frizzo Stefenon, Gurmail Singh, Kin-Choong Yow, and Alessandro Cimatti,
\newblock ``Semi-protopnet deep neural network for the classification of
  defective power grid distribution structures,''
\newblock {\em Sensors}, vol. 22, no. 13, pp. 4859, 2022.

\bibitem{singh2021these}
Gurmail Singh and Kin-Choong Yow,
\newblock ``These do not look like those: An interpretable deep learning model
  for image recognition,''
\newblock {\em IEEE Access}, vol. 9, pp. 41482--41493, 2021.

\bibitem{zhang2022protgnn}
Zaixi Zhang, Qi~Liu, Hao Wang, Chengqiang Lu, and Cheekong Lee,
\newblock ``Protgnn: Towards self-explaining graph neural networks,''
\newblock in {\em Proceedings of the AAAI Conference on Artificial
  Intelligence}, 2022, vol.~36, pp. 9127--9135.

\bibitem{kim2022vit}
Sangwon Kim, Jaeyeal Nam, and Byoung~Chul Ko,
\newblock ``Vit-net: Interpretable vision transformers with neural tree
  decoder,''
\newblock in {\em International Conference on Machine Learning}. PMLR, 2022,
  pp. 11162--11172.

\bibitem{hase2019interpretable}
Peter Hase, Chaofan Chen, Oscar Li, and Cynthia Rudin,
\newblock ``Interpretable image recognition with hierarchical prototypes,''
\newblock in {\em Proceedings of the AAAI Conference on Human Computation and
  Crowdsourcing}, 2019, vol.~7, pp. 32--40.

\bibitem{donnelly2022deformable}
Jon Donnelly, Alina~Jade Barnett, and Chaofan Chen,
\newblock ``Deformable protopnet: An interpretable image classifier using
  deformable prototypes,''
\newblock in {\em Proceedings of the IEEE/CVF Conference on Computer Vision and
  Pattern Recognition}, 2022, pp. 10265--10275.

\bibitem{dong_few-shot_2018}
Nanqing Dong and Eric~P Xing,
\newblock ``Few-shot semantic segmentation with prototype learning.,''
\newblock in {\em {BMVC}}, 2018, vol.~3.

\bibitem{ke2021prototypical}
Lei Ke, Xia Li, Martin Danelljan, Yu-Wing Tai, Chi-Keung Tang, and Fisher Yu,
\newblock ``Prototypical cross-attention networks for multiple object tracking
  and segmentation,''
\newblock {\em Advances in Neural Information Processing Systems}, vol. 34, pp.
  1192--1203, 2021.

\bibitem{gepperth2021gradient}
Alexander Gepperth and Benedikt Pf{\"u}lb,
\newblock ``Gradient-based training of gaussian mixture models for
  high-dimensional streaming data,''
\newblock {\em Neural Processing Letters}, vol. 53, no. 6, pp. 4331--4348,
  2021.

\bibitem{gautam2023looks}
Srishti Gautam, Marina M-C H{\"o}hne, Stine Hansen, Robert Jenssen, and Michael
  Kampffmeyer,
\newblock ``This looks more like that: Enhancing self-explaining models by
  prototypical relevance propagation,''
\newblock {\em Pattern Recognition}, vol. 136, pp. 109172, 2023.

\bibitem{xiong2010semi}
Biao Xiong, XiaoJun Zhang, and WanShou Jiang,
\newblock ``Semi-supervised classification based on gaussian mixture model for
  remote imagery,''
\newblock {\em Science China Technological Sciences}, vol. 53, no. 1, pp.
  85--90, 2010.

\bibitem{girshick2015fast}
Ross Girshick,
\newblock ``Fast r-cnn,''
\newblock in {\em Proceedings of the IEEE international conference on computer
  vision}, 2015, pp. 1440--1448.

\bibitem{achanta2010slic}
Radhakrishna Achanta, Appu Shaji, Kevin Smith, Aurelien Lucchi, Pascal Fua, and
  Sabine S{\"u}sstrunk,
\newblock ``Slic superpixels,''
\newblock Tech. {R}ep., 2010.

\bibitem{he2017mask}
Kaiming He, Georgia Gkioxari, Piotr Doll{\'a}r, and Ross Girshick,
\newblock ``Mask r-cnn,''
\newblock in {\em Proceedings of the IEEE international conference on computer
  vision}, 2017, pp. 2961--2969.

\bibitem{hani2019minneapple}
Nicolai Häni, Pravakar Roy, and Volkan Isler,
\newblock ``Minneapple: A benchmark dataset for apple detection and
  segmentation,'' 2019.

\bibitem{santos2020grape}
Thiago~T Santos, Leonardo~L de~Souza, Andreza~A dos Santos, and Sandra Avila,
\newblock ``Grape detection, segmentation, and tracking using deep neural
  networks and three-dimensional association,''
\newblock {\em Computers and Electronics in Agriculture}, vol. 170, pp. 105247,
  2020.

\bibitem{hani2020minneapple}
Nicolai H{\"a}ni, Pravakar Roy, and Volkan Isler,
\newblock ``Minneapple: a benchmark dataset for apple detection and
  segmentation,''
\newblock {\em IEEE Robotics and Automation Letters}, vol. 5, no. 2, pp.
  852--858, 2020.

\end{thebibliography}
\end{document}